\documentclass[conference, letterpaper, 10pt]{IEEEtran}
\pdfoutput=1
\usepackage[inline]{enumitem}
\usepackage{hyperref}
\input{config.sty}
\acrodef{CA}{Certification Authority}
\acrodef{TLS}{Transport Layer Security}
\acrodef{DT}{Digital Twin}
\acrodef{MI}{Mutual Information}
\acrodef{CE}{Cross-Entropy}
\acrodef{FE}{Feature Engineering}
\acrodef{MEC}{Multi-Access Edge Computing} 
\acrodef{mMTC}{massive Machine-Type Communications}
\acrodef{NTN}{Non Terrestrial Network}
\acrodef{PLR}{Packet Loss Rate}
\acrodef{PRR}{Packet Received Ratio}
\acrodef{BER}{Bit Error Rate}
\acrodef{ITS}{Intelligent Transport system}
\acrodef{RL}{Reinforcement Learning}
\acrodef{NN}{Neural Network}
\acrodef{ADAS}{Advanced Driver-Assistance Systems}
\acrodef{RTT}{Round-Trip Time}
\acrodef{HAP}{High Altitude Platform}
\acrodef{SC}{Streaming Client}
\acrodef{IMU}{Inertial Measurement Unit}
\acrodef{AI}{Artificial Intelligence}
\acrodef{AIaaS}{AI-as-a-Service}
\acrodef{CNN}{Convolutionl Neural Network}
\acrodefplural{UW2C}{user-wide-wearable-collections} 
\acrodef{OEC}{Orbital Edge Computing}
\acrodef{TCP}{Transmission Control Protocol}
\acrodef{MCS}{Mobile Crowd-Sensing}
\acrodef{UDP}{User Datagram Protocol}
\acrodef{AIMD}{Additive Increase Multiplicative Decrease}
\acrodef{HOG}{Histogram of Oriented Gradient}
\acrodef{FAS}{Federated Aerospace System}
\acrodef{CPSoS}{Cyber-Physical System of Systems}
\acrodef{CPS}{Cyber Physical System}
\acrodef{PEP}{Performance Enhancing Proxy}
\acrodef{CWND}{congestion window}
\acrodef{STP}{Satellite Transport Protocol}
\acrodef{HSMS}{Human State Monitoring System}
\acrodef{BDP}{Bandwidth-Delay Product}
\acrodef{QoS}{Quality of Service}
\acrodef{LoS}{Line of Sight}
\acrodef{NLoS}{Non Line of Sight}
\acrodef{BLoS}{Beyond Line of Sight}
\acrodef{QoE}{Quality of Experience}
\acrodef{ACM}{Adaptive Coding and Modulation}
\acrodef{DAMA}{Demand Assignment Multiple Access}
\acrodef{VANET}{Vehicular Ad-Hoc Network}
\acrodef{RA}{Random Access}
\acrodef{DA}{Dedicated Access}
\acrodef{OS}{Operating System}
\acrodef{CRA}{Contention Resolution ALOHA}
\acrodef{SA}{Slotted ALOHA}
\acrodef{DSA}{Diversity Slotted ALOHA}
\acrodef{OBU}{On-Board Unit}
\acrodef{CRDSA}{Contention Resolution Diversity Slotted ALOHA}
\acrodef{SIC}{Successive Interference Cancellation}
\acrodef{IC}{Interference Cancellation}
\acrodef{SINR}{Signal-to-Interference-plus-Noise Ratio}
\acrodef{ARQ}{Automatic Repeat reQuest}
\acrodef{SC-ARQ}{Selective-Coded ARQ}
\acrodef{SR-ARQ}{Selective-Repeat ARQ}
\acrodef{IRSA}{Irregular Repetition Slotted ALOHA}
\acrodef{CGC}{Complementary Ground Component}
\acrodef{RSU}{Road-Side Unit}
\acrodef{RSI}{Road-Side Infrastructure}
\acrodef{ACK}{Acknowledgment}
\acrodef{S-NS3}{Satellite Network Simulator}
\acrodef{PMF}{Probability Mass Function}
\acrodef{NACK}{Negative Acknowledgment}
\acrodef{DVB-SH}{Digital Video Broadcasting - Satellite Services to Handhelds}
\acrodef{DVB-H}{Digital Video Broadcasting - Handheld}
\acrodef{DVB-RCS2}{Digital Video Broadcasting - Return Channel via Satellite}
\acrodef{SACK}{Selective Acknowledgment}
\acrodef{SNACK}{Selective Negative Acknowledgment}\acrodef{SNACK}{Selective Negative Acknowledgment}
\acrodef{SNIR}{Signal to Noise plus Interference Ratio}
\acrodef{SCPS-TP}{Space Communications Protocol Specifications - Transport Protocol}
\acrodef{CCSDS}{Consultative Committee for Space Data Systems}
\acrodef{ESA}{European Space Agency}
\acrodef{NASA}{National Aeronautics and Space Administration}
\acrodef{BSM}{Broadband Satellite Multimedia}
\acrodef{RLNC}{Random Linear Network Coding}
\acrodef{NC}{Network Coding}
\acrodef{FIFO}{First In, First Out}
\acrodef{FCFS}{First Come, First Served}
\acrodef{BLER}{Block Error Rate}
\acrodef{GEO}{Geosynchronous}
\acrodef{LEO}{Low-Earth Orbit}
\acrodef{FTP}{File Transfer Protocol}
\acrodef{CRC}{Cyclic Redundancy Check}
\acrodef{MAC}{Media Access Control}
\acrodef{HTTP}{Hypertext Transfer Protocol}
\acrodef{ISP}{Internet Service Provider}
\acrodef{MSS}{Maximum Segment Size}
\acrodef{BIC}{Binary Increase Congestion control}
\acrodef{AQM}{Active Queue Management}
\acrodef{XCP}{eXplicit Control Protocol}
\acrodef{ECN}{Explicit Congestion Notification}
\acrodef{RED}{Random Early Detection}
\acrodef{TD}{Triple-Duplicate}
\acrodef{TO}{TimeOut}
\acrodef{MEO}{Medium-Earth Orbit}
\acrodef{NR}{new-radio}
\acrodef{IP}{Internet Protocol}
\acrodef{WMN}{Wireless Mesh Network}
\acrodef{ssthresh}{Slow-Start threshold}
\acrodef{MPE-IFEC}{Multi Protocol Encapsulation - Inter-burst Forward Error Correction}
\acrodef{FEC}{Forward Error Correction}
\acrodef{FSA}{Framed Slotted ALOHA}
\acrodef{D-FSA}{Diversity - Framed Slotted ALOHA}
\acrodef{CSA}{Coded Slotted ALOHA}
\acrodef{ML}{Machine Learning}
\acrodef{CRC}{Cyclic Redundancy Check}
\acrodef{P2P}{Peer-to-Peer}
\acrodef{EIRP}{Effective Isotropic Radiated Power}
\acrodef{FMT}{Fade Mitigation Technique}
\acrodef{SGD}{Smart Gateway Diversity}
\acrodef{NCC}{Network Control Centre}
\acrodef{ModCod}{Modulation and Coding}
\acrodef{FIFO}{First-In-First-Out}
\acrodef{WRR}{Weighted Round Robin}
\acrodef{WFQ}{Weighted Fair Queuing}
\acrodef{NS}{Network Simulator}
\acrodef{GSE}{Generic Stream Encapsulation}
\acrodef{PDF}{Probability Density Function}
\acrodef{CDF}{Cumulative Density Function}
\acrodef{AWGN}{Additive White Gaussian Noise}
\acrodef{CoV}{Coefficient of Variation}
\acrodef{MSC}{Message Sequence Chart}
\acrodef{ESA}{European Space Agency}
\acrodef{LIU}{Lebanese International University}
\acrodef{TUM}{Technical University of Munich}
\acrodef{MSCE-CS}{Master of Science in Communications Engineering - Communications Systems}
\acrodef{DLR}{German Aerospace Center}
\acrodef{NC-SGD}{Network Coding for SGD}
\acrodef{ATSP}{Advanced Transport Satellite Protocol}
\acrodef{PoI}{packet of interest}
\acrodef{STP}{Satellite Transport Protocol}
\acrodef{WMN}{Wireless Mesh Network}
\acrodef{SNR}{Signal-to-Noise Ratio}
\acrodef{SINR}{Signal-to-Interference-plus-Noise Ratio}
\acrodef{LMS}{Land Mobile Satellite}
\acrodef{LTE}{Long-Term Evolution}
\acrodef{M2M}{Machine-to-Machine}
\acrodef{IoT}{Internet of Things}
\acrodef{RA}{Random Access}
\acrodef{UAV}{Unmanned Aerial Vehicle}
\acrodef{UAS}{Unmanned Aerial System}
\acrodef{FANET}{Flying Ad-Hoc Network}
\acrodef{MANET}{Mobile Ad-Hoc Network}
\acrodef{VANET}{Vehicle Ad-Hoc Network}
\acrodef{C2}{Command and Control}
\acrodef{DTN}{Delay Tolerant Network}
\acrodef{COTS}{Commercial Off-the-Shelf}
\acrodef{IETF}{Internet Engineering Task Force}
\acrodef{CoAP}{Constrained Application Protocol}
\acrodef{MQTT}{Message Queue Telemetry Transport}
\acrodef{CoRE}{Constrained RESTful Environments}
\acrodef{ROLL}{Routing Over Low power and Lossy networks}
\acrodef{6Lo}{IPv6 over Networks of Resource-constrained Nodes}
\acrodef{URI}{Uniform Resource Identifier}
\acrodef{PUB/SUB}{Publish / Subscribe}
\acrodef{RCST}{Return Channel Satellite Terminal}
\acrodef{TDMA}{Time Division Multiple Access}
\acrodef{FDMA}{Frequency Division Multiple Access}
\acrodef{TCDMA}{Turbo Code Division Multiple Access}
\acrodef{PDMA}{Power Division Multiple Access}
\acrodef{WSN}{Wireless Sensor Network}
\acrodef{REST}{Representational State Transfer}
\acrodef{EDGE}{Enhanced Data rates for GSM Evolution}
\acrodef{UMTS}{Universal Mobile Telecommunications System}
\acrodef{LTE}{Long-Term Evolution}
\acrodef{E2E}{End-to-End}
\acrodef{3WHS}{Three-way Handshake}
\acrodef{SCADA}{Supervisory Control And Data Acquisition}
\acrodef{SOA}{Service-Oriented Architecture}
\acrodef{6LoWPAN}{IPv6 over Low power Wireless Personal Area Networks}
\acrodef{CoCoA}{CoAP Simple Congestion Control/Advanced}
\acrodef{RTO}{Retransmission TimeOut}
\acrodef{GPRS}{General Packet Radio Service}
\acrodef{TFRC}{TCP Friendly Rate Control}
\acrodef{DTMC}{Discrete Time Markov Chain}

\acrodefplural{RSU}[RSUs]{Road Side Units}
\acrodefplural{VANET}[VANETs]{Vehicular Ad-Hoc Networks}
\acrodefplural{OBU}[OBUs]{On-Board Units}
\acrodefplural{UAV}[UAVs]{Unmanned Aerial Vehicles}
\acrodefplural{FANET}[FANETs]{Flying Ad-Hoc Networks}
\acrodefplural{MANET}[MANETs]{Mobile Ad-Hoc Networks}
\acrodefplural{VANET}[VANETs]{Vehicle Ad-Hoc Networks}
\acrodefplural{UAS}[UASs]{Unmanned Aerial Systems}
\acrodefplural{RCST}[RCSTs]{Return Channel Satellite Terminal}

\title{AI-as-a-Service Toolkit for Human-Centered Intelligence in Autonomous Driving}

\DeclareRobustCommand*{\IEEEauthorrefmark}[1]{%
  \raisebox{0pt}[0pt][0pt]{\textsuperscript{\footnotesize\ensuremath{#1}}}}

\author{
\IEEEauthorblockN{
Valerio De Caro\IEEEauthorrefmark{1}\textsuperscript{\textsection}, 
Saira Bano\IEEEauthorrefmark{2}\textsuperscript{,}\IEEEauthorrefmark{3}\textsuperscript{\textsection}, 
Achilles Machumilane\IEEEauthorrefmark{2}\textsuperscript{,}\IEEEauthorrefmark{3}\textsuperscript{\textsection}, 
Alberto Gotta\IEEEauthorrefmark{3}, 
Pietro Cassarà\IEEEauthorrefmark{3}, \\
Antonio Carta\IEEEauthorrefmark{1}, 
Rudy Semola\IEEEauthorrefmark{1}, 
Christos Sardianos\IEEEauthorrefmark{4}, 
Christos Chronis\IEEEauthorrefmark{4}, 
Iraklis Varlamis\IEEEauthorrefmark{4}, \\
Konstantinos Tserpes\IEEEauthorrefmark{4},
Vincenzo Lomonaco\IEEEauthorrefmark{1}, 
Claudio Gallicchio\IEEEauthorrefmark{1} 
and Davide Bacciu\IEEEauthorrefmark{1}}
\IEEEauthorblockA{
\IEEEauthorrefmark{1}Department of Computer Science, University of Pisa, Pisa, Italy \\
\IEEEauthorrefmark{2}Department of Information Engineering, University of Pisa, Pisa, Italy \\
\IEEEauthorrefmark{3}Information Science and Technology Institute "A. Faedo", National Research Council, Pisa, Italy \\
\IEEEauthorrefmark{4}Harokopio University of Athens, Athens, Greece
}
}

\begin{document}

\maketitle
\begingroup\renewcommand\thefootnote{\textsection}
\footnotetext{Equal contribution}
\endgroup

\begin{abstract}
This paper presents a proof-of-concept implementation of the AI-as-a-Service toolkit developed within the H2020 TEACHING project and designed to implement an autonomous driving personalization system according to the output of an automatic driver’s stress recognition algorithm, both of them realizing a Cyber-Physical System of Systems. In addition, we implemented a data-gathering subsystem to collect data from different sensors, i.e., wearables and cameras, to automatize stress recognition. 
The system was attached for testing to a driving emulation software, CARLA, which allows testing the approach's feasibility with minimum cost and without putting at risk drivers and passengers.
At the core of the relative subsystems, different learning algorithms were implemented using Deep Neural Networks, Recurrent Neural Networks, and Reinforcement Learning.
\end{abstract}

\section{Introduction \& Motivation}
The TEACHING project\footnote{\href{https://teaching-h2020.eu}{https://teaching-h2020.eu}} aims at designing a computing platform and the associated software toolkit to support the development and the deployment of adaptive and dependable \ac{CPSoS} applications, enabling them to exploit sustainable human feedback to drive, optimize and personalize the provisioning of the offered services.
The concept of Humanistic Intelligence in autonomous \ac{CPSoS} is central to the project, where the human and the cybernetic components cooperate in the process of mutual empowerment. The concept of distributed, efficient, and dependable \ac{AI}, leveraging edge computing support, is hence fundamental to achieving the goals of the project. 
One of the essential goals of the project is to design methodologies and tools to create the TEACHING \ac{AIaaS} software infrastructure for \ac{CPSoS} applications, developing human-centric personalization mechanisms \cite{bacciu2021teaching}.
To this aim, the activities were organized in order to \textit{i)} develop a toolkit that implements the core methodological components of the TEACHING \ac{AIaaS}; \textit{ii)} develop AI methodologies that are specialized in the recognition and characterization of the human physiological, emotional, and cognitive state from streams of data gathered from heterogeneous sensors; \textit{iii)} develop the necessary functionalities to self-adapt and personalize \ac{AI} models of autonomous driving, using the human state information implicitly; \textit{iiii)} develop privacy-aware AI methodologies that can be bundled within the AIaaS toolkit.

In this work, we present a demonstrator which showcases the first three points in an autonomous driving simulation. In particular, the user experiences an autonomous driving simulation, while being monitored by a wearable sensor and a smart camera. The data collected by these devices are gathered into the AIaaS platform, which employs two main learning modules: one for detecting the cognitive state of the user from the physiological data, and the other for personalizing the driving style to one that better suits the state of the user. Data from both sensors and learning modules are then monitored and plotted by an additional application delegated to such purpose.

\section{The AIaaS Toolkit}

\begin{figure}[t]
    \centering
    \includegraphics[width=0.8\columnwidth]{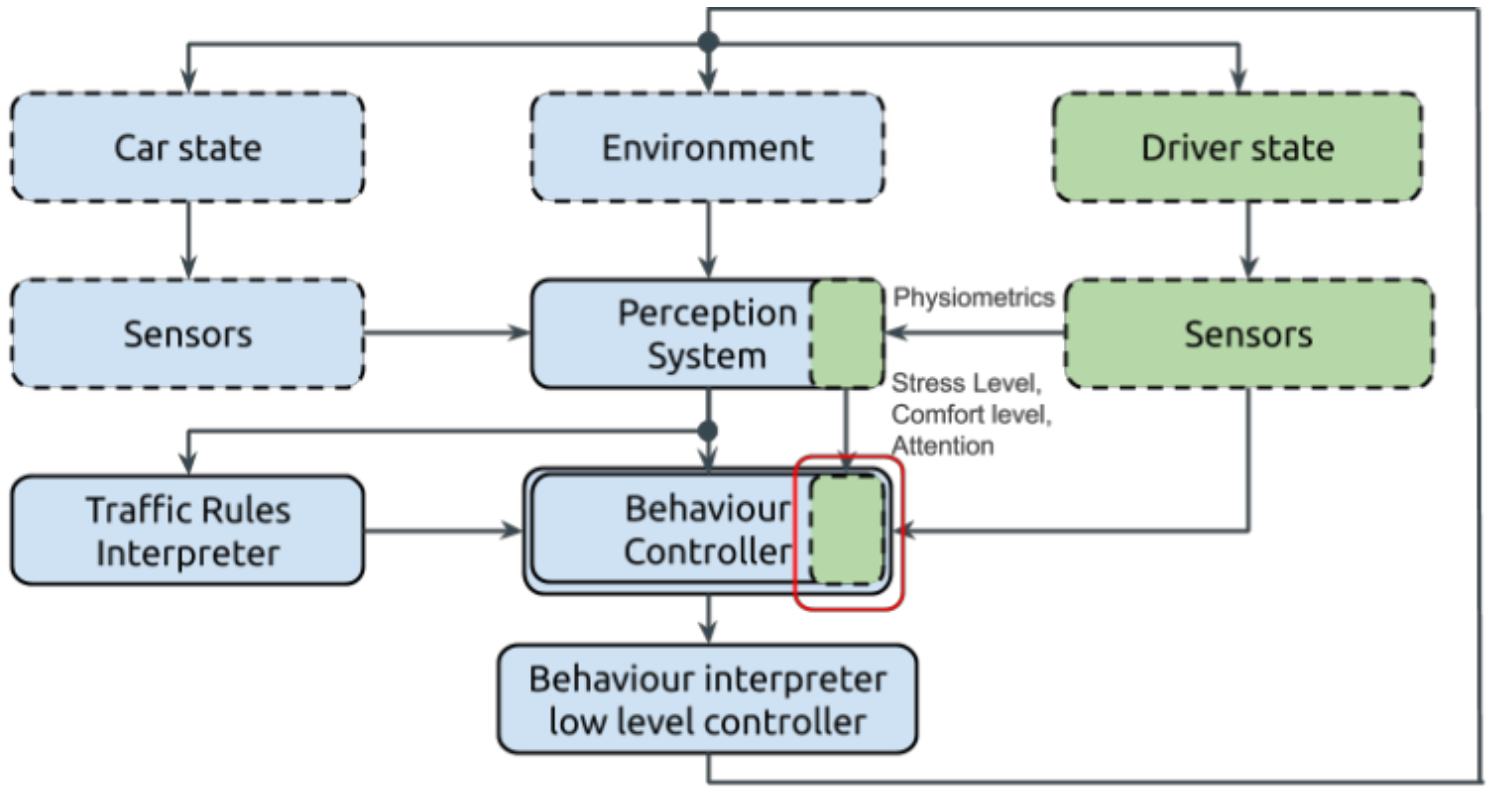}
    \caption{Overview of the presented control system.}
    \label{fig:system}
\end{figure}

The cognitive state of a human in a driving experience depends on the perception of the vehicle condition and the state of its environment, as well as on the internal beliefs, preferences, and expectations from an autonomous driving scenario. Several studies on perception and the consequent mental state \cite{brookhuis2010monitoring, begum2012mental} fuse data from a large variety of sensors attached to the vehicle (e.g., accelerometers, gyroscopes, etc.) and the driver in an attempt to detect the driver's driving style and associate it to the driver/passengers' comfort or stress level \cite{bellem2016objective}.

Human-centered intelligence in autonomous driving can be achieved by adapting the driving behaviour according the corresponding contextual data. As sketched in Figure \ref{fig:system}, such data can be categorized in three main types, depending on their availability along the driving experience:
\begin{enumerate}
    \item \textit{Vehicle's state}, which is monitored with the use of sensors attached to the vehicle, the driver and the passengers;
    \item \textit{Environment's state}, which denotes the outward context that can be used to achieve awareness of the situation occurring at any moment;
    \item \textit{Driver's state}, which can be classified by acquiring physiological signals through inward monitoring cameras or physiological sensors embedded in smart wearable devices.
\end{enumerate}
In this context, the adaptation of the driving behaviour consists in performing actions to modify the vehicle's behaviour in order to provide a driving profile which better suites the cognitive state of the driver. 

It is clear that the complexity of the task can scale at will, depending on the different HW/SW setups, as well as the different modelling choices from the \ac{AI} perspective, making the definition of the dedicated software arbitrarily complex. The AIaaS toolkit that we show in this paper aims to streamline the definition of dedicated software by relying on a set of composable, hardware-agnostic app-building blocks called Learning Modules (LM) and data sources, which allow to design reusable and portable AI applications. In particular, the LM building blocks can be separately ported to and optimized for different device HW/SW architectures, increasing their efficiency with respect to common metrics (performance, power consumption), allowing careful debugging and verification, as well as allowing to exploit specific features of the execution platform within the LM. Communications between the modules of the AIaaS toolkit happen in two main ways, depending on whether the interactions are internal (i.e., between modules deployed within the vehicle) or external (i.e., from a model deployed within the vehicle, towards an external one). For the former, we employed a data brokering approach, in which publishers publish their own data on the message broker, while subscribers can gather specific data from it. For the latter, we employed a distributed event streaming platform, which enables the toolkit to high-performance communications towards edge and cloud. Finally, the AIaaS toolkit enables to humanistic intelligence, since, with an appropriate interaction between the modules, it is possible to produce a loop in which the human is implicitly an integral part.

\section{Demo Description}
\subsection{Modelling of the Scenario}
As mentioned previously, the autonomous driving scenario is subject to a wide variety of tasks, depending on either the HW/SW support and the modelling choices of the driver's state, the vehicle's state and the environment's state. In this work, we showcase a simulated scenario in which the task consists on choosing the best driving style, depending only on the driver's state. Notice that the perception of the vehicle and the environment from the human affects their cognitive state, thus closing the loop with the system.
In particular, the driver's state is modelled by two main components:
\begin{itemize}
    \item \textit{Facial Action Coding System} (FACS), which is a system for classifying facial movements as they appear in various human expressions \cite{FACS}. Such movements of individual facial muscles are coded by FACS attributing a combination of codes corresponding to certain micro-movements, called AUs, performed by the person. FACS can systematically encode almost all anatomically possible facial expressions, reducing them to a specific subset of AUs;
    \item \textit{Cognitive state} of the driver, which we simplified with the notion of \textit{cognitive stress}. Several papers in the literature employ physiological signals for identifying drivers' cognitive state \cite{brookhuis2010monitoring} and comfort levels \cite{bellem2016objective} under varying driving conditions. 
\end{itemize}
Instead, the adaptation of the driving style consists in leveraging on predictions on the driver's state to predict the best driving profile for the vehicle's \ac{ADAS} from a predefined set. In the context, the task of personalizing the driving style is typically performed with the use of a model that can "learn from humans" how a real human would adjust the car's controls in various conditions. Recently, authors in \cite{kendall2019learning} designed a deep \ac{RL} algorithm and used the deterministic policy gradients algorithm for training an asynchronous Advantage Actor-Critic (A3C) \ac{RL} agent to maximize a reward that corresponds to the distance travelled by the vehicle without human interference. Nevertheless, most models are based on an expected average conscious user, and their control parameters are predefined and cannot be altered to match the different driver profiles. Thus, in our setting, we opted for a different approach to learn only with the use of a data coming from sensors and simulated scenarios.

\subsection{Design and Components}
\begin{figure}[t]
    \centering
    \includegraphics[width=0.90\columnwidth, keepaspectratio]{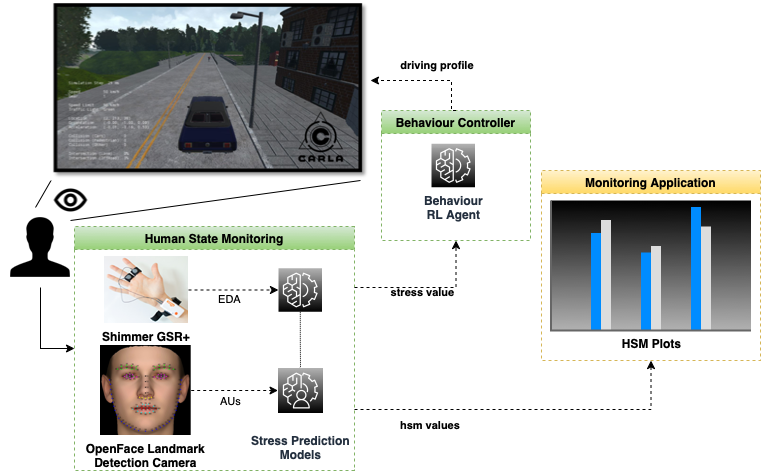}
    \caption{Overview of the demonstrator. Dashed arrows represent communications via MQTT.}
    \label{fig:demo}
\end{figure}
In Figure \ref{fig:demo}, the demo platform is briefly presented: at the top of the figure, CARLA emulator \cite{Carla} is the autonomous driving simulator, which provides the interaction with the driver. Data extracted by visual and physiological sensors provide the input to a ML model that yields the driver's stress level. According to this output, a \ac{RL} algorithm provides a set of actions that personalize the driving profile simulated in CARLA aiming to resolve the driver's stress level. In the following, we will discuss the details of each of the interacting components of our demo.

\subsubsection{Physiological Data} these data were collected with the use of a Shimmer3 GSR+ Unit, which monitors galvanic skin response between two reusable electrodes attached to the two fingers of one hand. A stimulus can activate the sweat glands, increasing moisture on the skin and allowing the current to flow more readily by changing the balance of positive and negative ions in the secreted fluid (increasing skin conductance). The Shimmer3 GSR+ Unit can also provide a PPG signal from a finger, ear lobe, or other body part by adding an Optical Pulse probe. This signal is used as a proxy to compute the Heart Rate (HR) of the individual, which, in this demo, we used for a monitoring purpose only.

\subsubsection{Visual Data}
\begin{figure}[t]
    \centering
    \includegraphics[width=.45\columnwidth]{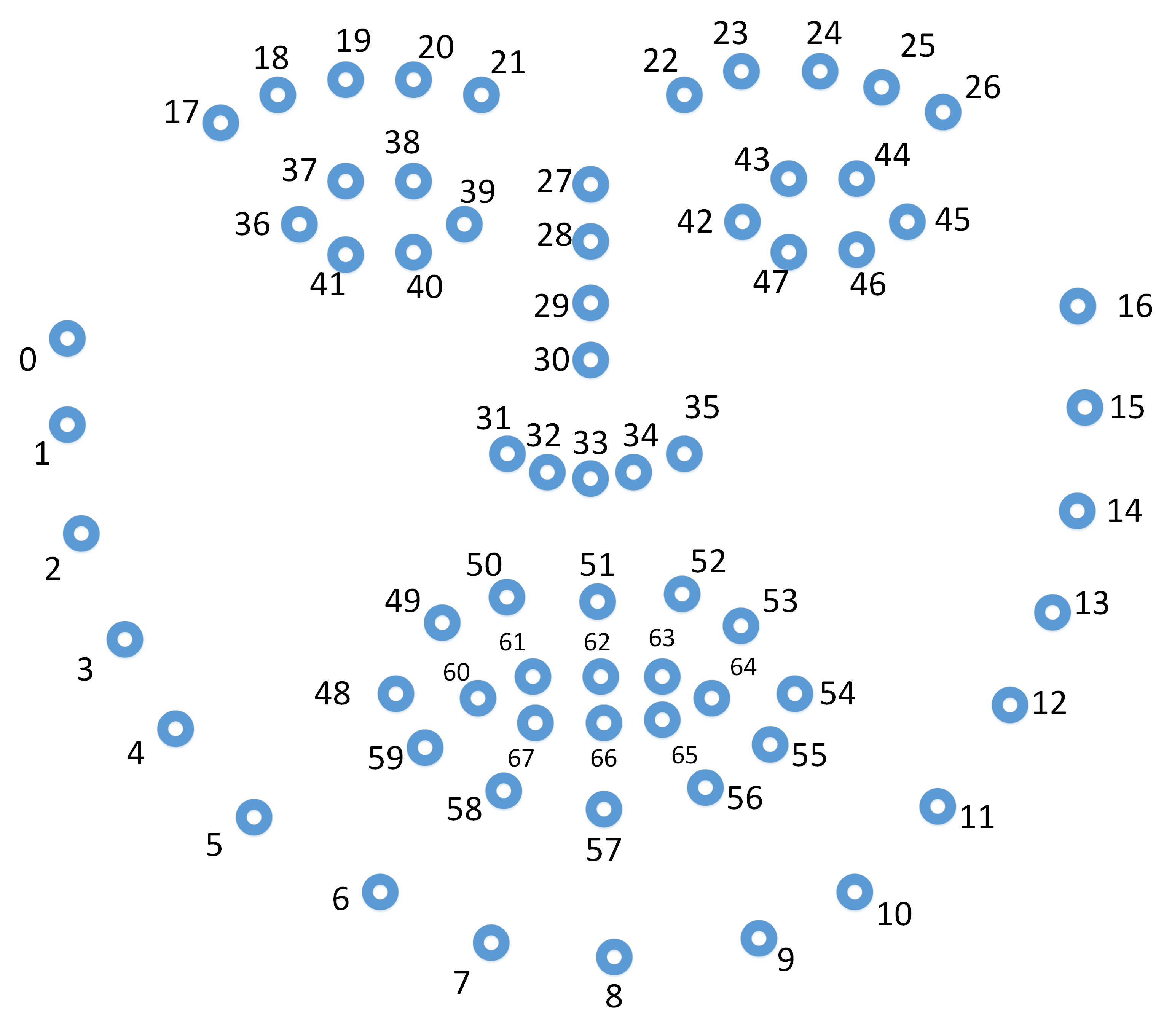}
    \caption{The 68 landmarks mapped onto the face of the subject.}
    \label{fig:landmark}
\end{figure}
for collecting and processing these data, we employed OpenFace, which is an open-source software developed for computer vision applications based on facial behaviour analysis \cite{openface}. Such analysis is performed by detecting 68 \textit{facial landmark locations} (Figure \ref{fig:landmark}), which are employed for understanding emotional states; \textit{head pose}, which is related to the expression of emotions and social cues; \textit{gaze direction}, which is crucial when assessing attention, mental health, and emotion intensity. With the use of dedicated learning algorithms \cite{openface}, Openface can determine the features mentioned above by analyzing faces in video files, in computer images or even through a live camera, as for this demo, without any specialized hardware.

\subsubsection{Data-Gathering and Storage} sensors are accessed through a common, dedicated Sensors API, but not directly, in order to provide a place for adapter components that insulate the application support from the specifics of different kind of sensors. The data is then collected by the Data Brokering component, which takes the role of main distributor of data inside the AIaaS platform through a publish/subscribe messaging paradigm, allowing the selection of data of interest to each client.
To this aim,  we employed the MQTT protocol, and instantiated it with a Mosquitto\footnote{\href{https://mosquitto.org}{https://mosquitto.org}} broker. Data from the physiological sensor were gathered thanks to its dedicated SDK, while an interface was developed to integrate OpenFace into the data-gathering subsystem.

\subsubsection{Stress Detection} the goal of this module is to recognize the level of stress from physiological sensors. For this demo, we monitored the EDA to provide a continuous measure of the current level of stress through the inference of an Echo State Network (ESN), trained over a public dataset (WESAD). ESNs are models which belong to the family of recurrent neural networks, and are known for their efficiency in both training and inference, which makes them suitable for their use in embedded systems. This module showcases how trivial it is, with the \ac{AIaaS} framework, to insert the human in the loop by adding dedicated modules (i.e., sensors and Learning Modules) to the application.

\subsubsection{Driving Style Personalization}
an \ac{RL} method implements an Actor-Critic algorithm to train an agent that chooses driving profile which best suits the cognitive state of the driver \cite{chronis2021pci}. More specifically, the value network predicts the driver's stress values with respect to the vehicle's behaviour, while a policy network outputs an action for CARLA's agent. The action consists in  choosing between three driving profiles, i.e., conservative, normal and aggressive, which differ in terms of driving parameters like maximum speed, maximum steering angle and acceleration. These profiles are implemented with the use of the CARLA API, which allows to set the behaviour of the vehicle and the environment, as well as collecting the data from simulated sensors along the simulation.

\section{Conclusion}
In this paper we present a proof-of-concept of the AI-as-a-Service toolkit, developed within the H2020 TEACHING project. We showcase an autonomous driving scenario in which a driving personalization system predicts a driving profile according to the driver's stress. In particular, a learning algorithm evaluates the driver's stress level with respect to the physiological data captured by a wearable sensor, while a \ac{RL} agent adapts CARLA's driving profile to reduce the driver's stress. Furthermore, the driver is monitored by a smart camera, which extracts features about the facial behaviour. 
The demo, consisting in a driving session simulated through the CARLA software, highlights how easily the AIaaS toolkit allows to insert humanistic intelligence in AI applications.

\ifCLASSOPTIONcompsoc
  \section*{Acknowledgments}
\else
  \section*{Acknowledgment}
\fi

\thanks{This work is supported by the TEACHING project funded by the EU Horizon 2020 under GA n. 871385}



\bibliographystyle{IEEEtran}

\bibliography{references}

%



\end{document}